\documentclass[review]{elsarticle}

\usepackage{lineno,hyperref,amsmath}
\usepackage{diagbox}
\usepackage{subfigure}

\modulolinenumbers[5]

\journal{Journal of \LaTeX\ Templates}

\begin{document}

\begin{frontmatter}

\title{MG-SAGC: A multiscale graph and its self-adaptive graph convolution network for 3D point clouds}

\author[SHR,SA]{Bo Wu}
\ead{wubo@nlsde.buaa.edu.cn}

\author[SHR]{Bo Lang}
\ead{langbo@buaa.edu.cn}

\address[SHR]{Beihang University, Beijing, China,}
\address[SA]{Purple Mountain Lab, Nanjing, China}






\begin{abstract}
To enhance the ability of neural networks to extract local point cloud features and improve their quality, in this paper, we propose a multiscale graph generation method and a self-adaptive graph convolution method. First, we propose a multiscale graph generation method for point clouds. This approach transforms point clouds into a structured multiscale graph form that supports multiscale analysis of point clouds in the scale space and can obtain the dimensional features of point cloud data at different scales, thus making it easier to obtain the best point cloud features. Because traditional convolutional neural networks are not applicable to graph data with irregular vertex neighborhoods, this paper presents an sef-adaptive graph convolution kernel that uses the Chebyshev polynomial to fit an irregular convolution filter based on the theory of optimal approximation. In this paper, we adopt max pooling to synthesize the features of different scale maps and generate the point cloud features. In experiments conducted on three widely used public datasets, the proposed method significantly outperforms other state-of-the-art models, demonstrating its effectiveness and generalizability.
\end{abstract}

\begin{keyword}
\texttt multiscale graph \sep self-adaptive graph convolution \sep Chebyshev polynomial \sep point clouds
\MSC[2010] 00-01\sep  99-00
\end{keyword}

\end{frontmatter}


\section{Introduction}

In 3-dimensional (3D) point cloud classification and retrieval tasks, how to extract point cloud features is a popular issue. Previous researchers have conducted a series of studies on methods for extracting point cloud features. For example, Gumhold \cite{FEPC200143} et al. presented a method based on tensor voting theory to extract sharp features from a point cloud. Kim \cite{ERVL201344} et al. used scale voting theory to combine point clouds with K-mean clustering algorithms.Hsu \cite{EGFL200945} et al. proposed a feature extraction algorithm based on a triangular mesh. However, these existing point cloud feature extraction methods are designed for specific applications and the extracted features are designed for specific tasks. When facing a new task, these methods have difficulty achieving good effects, are difficult to optimize, and have insufficient generalization capabilities. In recent years, deep neural networks have performed well on classification and segmentation tasks for 2-dimensional (2D) images, which has inspired people to apply deep learning to point cloud data, where the original data are input to end-to-end deep learning methods. However, the properties of point cloud data introduce new challenges to deep learning. At present, deep learning networks require regular input data, and the data arrangement sequence affects network output. However, a point cloud is a set of discrete data that can be both irregular and unordered. Therefore, the network frameworks for 2D image classification cannot be applied directly to 3D point clouds.

\par
To address the problems mentioned above, researchers have developed a variety of neural network models for point clouds\cite{FEPC200143,EFS20111007,EFC20171007,FHDFU2010}; these mainly include multiview-based approaches and voxel-based approaches. The multiview approach extracts point cloud features by projecting multiple images of the 3D model and then using the processing methods applied to 2-dimensional images \cite{Jun2013Aerial,Manyun2018119,MVCNN2015119}. In contrast, the voxel-based methods convert point cloud data into volume representations \cite{GVCNN20181109,MVHBN20181109,VoxNet20151109} and then use a 3D CNN to learn the features. However, both the voxel-based and multiview-based approaches result in information loss. Moreover, the conversion from point cloud to voxels clearly increases the amount of 3D data, which can lead to performance degradations. Additionally, these two kinds of methods are not directly applicable to point cloud data. Consequently, Charles R. Qi \cite{PointNet20171007} et al. proposed the PointNet model, which can address point cloud data directly and adapt to the irregularity and disorder of point clouds, thus avoiding the problems of information loss and efficiency degradation. Since PointNet was proposed, extensive attention has been paid to the direct processing of point cloud data. However, the vertex features generated by PointNet consider only basic information, such as location and color, and ignore the relative spatial relationships between points. Overall, there is still considerable room for improving the features generated by PointNet.

\par
In view of the fact that graphs can fully express the shape and spatial structure information of point clouds, in this paper, we investigate and propose a feature extraction method for point clouds based on graphs. Point cloud data have the characteristics of discreteness and irregularity. When a point cloud is expressed as a graph structure, the problems to be solved include whether an edge exists between two vertices and how to set the weights of edges. The current common approach to this problem is to set a fixed threshold; then, when the distance between two vertices is less than or equal to the threshold, an edge exists between the vertices; otherwise, no edge exists. When the vertex radius is too small, there will be too few points in the neighborhood of some vertices, and when the vertex radius is too large, there will be too many points in the neighborhood; both conditions affect the extraction of local point cloud features. In this paper, inspired by the multiscale feature generation method applied to 2D images \cite{Scale20171007}, we present a 3D point cloud feature generation method based on a multiscale graph. By extracting features at multiple scales, multiscale analysis of point clouds in scale space can be carried out, allowing the optimal features of point clouds to be obtained more easily.

\par
In this paper, we first define the multiscale space concept for 3D point clouds and then propose the multiscale graph construction method. A multiscale graph is a graph composed by selecting different neighborhood radii for each vertex in a point cloud. Moreover, because the vertex neighborhood of the graph is not regularized grid data, the traditional convolution filter cannot be used to aggregate the information from the vertex neighborhoods. To address this problem, based on the function approximation theory, we propose a type of multiscale graph convolution filter by using the Chebyshev multinomial. Our methods can make good use of the local feature information in the point cloud, and the experimental results show that the quality of the point cloud feature is improved.
\par
The main contributions of this work are as follows:
\par
\paragraph{(1)}To the best of our knowledge, we are the first to propose a multiscale graphical representation of a 3D point cloud. A multiscale graph consists of graphical structured data composed using different neighborhood radii for a vertex. This type of graph reflects the structural properties of a point cloud at different scales and thus more comprehensively represents the point cloud characteristics.
\paragraph{(2)}We propose an adaptive graph convolution feature extraction method for point clouds. We design an adaptive graph convolution kernel based on the Chebyshev polynomial to dynamically generate the convolution parameters for the vertex neighborhood of the point cloud; then the convolution operation can be applied to the irregular vertex neighborhood to obtain the multiscale local features of the point cloud. Finally, high-level features are obtained by superimposing the feature extraction modules.

\section{Related work}
\par
Due to the irregular and unordered structure of 3D point cloud data, traditional deep learning methods cannot be directly applied to extract features from 3D point clouds. At present, the deep learning methods \cite{EFS20111007,CTSD20111007,BRAND20121007,F3DRP20101007,BEAFS20181007,JCMTL20171007,EFC20171007,FHDFU} for 3D point clouds can be divided into two main categories: those that convert point cloud data into some regular form and those that act directly on the original point cloud data\cite{OCTNET201638,CTFVP20171007,DDCOA2018,PLC2015,FNPCA2018,PPFNG2018,VMCNN201635,GDVM201636,FPFH20091007}.
\par

The most commonly used point cloud data regularization methods are the voxel mesh method\cite{EGFL200945,VoxNet20151109,GDVM201636,DCNOG2015} and the multiview method\cite{MVCNN2015119,GVCNN20181109}. In the voxel mesh approach, the mesh provides a regular structure, and mesh transformation solves the permutation problem. However, under this approach the large amount of data that must be transformed is a serious problem. For example, when adding a dimension to a 256*256 image results in a 256*256*256 image, which equates to 16,777,216 voxels. Even existing fast GPUs are slow to process this volume of data. Therefore, it is usually necessary to compromise and adopt a lower resolution; however, reducing the resolution can introduce quantization errors, resulting in less accurate feature extraction. Roman Klokov et al.\cite{JCMTL20171007} used a KD tree to create a point cloud with a certain ordered structure and learned the weight of each node in the tree, but this method was highly sensitive to rotation and noise. Based on the successful applications of CNNs with 2-dimensional images, Hao Su et al.\cite{MVCNN2015119} proposed the multiview convolutional neural network, called a multiview CNN (MVCNN). This model projects the 3D point cloud data into 2D space and then performs convolution operations on 2D images. Finally, image features from multiple angles are fused by the max pooling method. Yifan Feng \cite{GVCNN20181109} proposed the group-view convolutional neural network for 3D shape recognition (GVCNN), which is an improvement of the MVCNN in which the projected images are first grouped and weighted; then, the weighted images are averaged and pooled to obtain the final result. However, such methods often result in information loss compared to the raw point cloud data.

\par
In addition to the voxel and multiview methods mentioned above, other 3D point cloud regularization methods, such as spatial mapping, have also been proposed. Wohlkinger W et al.\cite{EFS20111007} proposed the Sparse Lattice Networks (SPLATNet) model, which maps point cloud data to a regularized data space and then operates on the data in the regularized space. By introducing a bilateral convolutional layer, Martin Kiefel et al. constructed a universal and flexible neural network structure called the sparse lattice network that was suitable for point cloud data. The core of this network is the bilateral convolutional layer. Yaoqing Yang et al.\cite{FNPCA2018} proposed the FoldingNet model, which reduced the point cloud dimension to 2D space to learn features and then mapped the features back to 3D space. Haowen Deng proposed a point-pair feature network (PPFNet) model that learns local descriptors geometrically along with highly perceivable global information.
\par
In recent years, some deep learning models that target irregular point clouds directly have emerged. Charles R. Qi et al.\cite{PointNet20171007} published the PointNet model in Computer Vision and Pattern Recognition (CVPR). PointNet was the first deep neural network to directly process unordered point cloud data. This model uses the original point cloud data without any preprocessing to perform deep learning. The author used the max pooling method to solve the rotation invariance problem of 3D objects through an affine transformation of the original point matrix. However, this method is weak with regard to local feature extraction. Later, Charles R. Qi published the PointNet++ model\cite{POINTNET201739} at the Neural Information Processing Systems (NIPS) conference. The model first conducted sampling and grouping for point clouds and then used the PointNet network to extract features from each small region. This method improved PointNet's ability to extracting local features from point cloud data, but it did not take full advantage of the correlations between points. Lu \cite{POINTSIFT200946} proposed a framework based on the PointSIFT operator that extended the traditional 2D image SIFT algorithm to the 3D point cloud domain and output a representation vector for each point cloud in end-to-end fashion. This framework encoded information in all directions and adaptively selected the appropriate representation scale.

\par
A point cloud can also be thought of as an irregular graph. In 2017, Martin Simonovsky\cite{Simonovsky2017Dynamic} transformed a point cloud dataset into a graph dataset in an experiment using the edge-conditioned filters in convolutional neural networks (ECC) model\cite{Simonovsky2017Dynamic}. The method used a multilayer perceptron (MLP) to fit the convolution filters. ECC performed better on multiple datasets than the voxel mesh-based method. However, the MLP was a fully connected structure with too many parameters to fit. The DGCNN model proposed a graph convolutional neural network method that used a Gaussian mixture model to fit the graph convolutional filter. However, this method needed to calculate a mixed Gaussian function once for each point, making its complexity relatively high.
\par
To solve the above problem, in this paper, we propose a self-adaptive graph convolution neural network method based on the Chebyshev polynomial. The Chebyshev polynomial with the best consistent approximation ability is used to fit the graph convolutional filter, which improves the model's fitting ability. In addition, we construct a deep neural model based on a multiscale neighborhood graph of a point cloud. The method takes advantage of the local information between the vertices in the 3D point cloud, which improves the feature expression ability of the point cloud.

\section{Method}
The existing point cloud feature generation networks usually extract features from each independent point or calculate global point cloud features based on single-scale neighborhood features of each point. Thus, these methods do not fully consider the mutual relations and structural constraints of local vertices in multiscale neighborhoods of the point cloud. In this paper, we propose a multiscale graph with stronger local expression ability; then, we design a self-adaptive convolution kernel for the graph to extract multiscale local features.
\par
The method proposed in this paper includes the following three main parts:
\par
\paragraph{(1)}Multiscale graph representation of point clouds. In the multiscale point cloud representation, because the density of vertices in the point cloud is inconsistent, we instead use the average point distance of the point cloud as the measurement baseline.
\paragraph{(2)}Efficient adaptive graph convolution. Due to the irregularity of the vertex neighborhoods in the point cloud, we design an adaptive convolution kernel based on the Chebyshev polynomial to aggregate the irregular vertex neighborhood information in a point cloud.
\paragraph{(3)}A deep neural network model based on the multiscale graph and the adaptive graph convolution. By combining the multiscale graph and the adaptive graph convolution kernel based on the Chebyshev polynomial, we construct a deep neural network model to extract point cloud features.

\subsection{Multiscale graph generation Method}
The multiscale graphs of the point cloud represent the structure of the point cloud under multiple scales. In this paper, we use an r-scale graph to represent a graph structure for a point cloud with radius $r$.
\par
\paragraph{Definition 1 r-scale graph:} For each point i in the original point cloud, a space s of radius r is established. If a point j lies within the space s adjacent to point $i$, then an edge is created between point $j$ and point $i$. The graph formed according to this rule is an $r$-scale graph.
\par
By adjusting the size of the scale radius $r$, we obtain different neighborhood sizes for point $i$; consequently, different $r$ values construct point cloud graphs with different scales.
\par
In an r-scale graph, any vertex $i$ and vertex $j$ in its neighborhood meet the following formula:
\begin{equation}
{{d}_{i}}_{j}\le r
\end{equation}
\begin{equation}
{{d}_{i}}_{j}=\sqrt{{{({{x}_{i}}-{{x}_{j}})}^{2}}+{{({{y}_{i}}-{{y}_{j}})}^{2}}+{{({{z}_{i}}-{{z}_{j}})}^{2}}}
\end{equation}
\par

where $d_{ij}$ represents the Euclidean distance between vertex $i$ and vertex $j$; $r$ represents the radius of the selected vertex; $x_i$, $y_i$, $z_i$ are the coordinates of vertex $i$; and $x_j$, $y_j$, $z_j$ are the coordinate of vertex $j$.

\par
\begin{figure}
\vspace{-0.1in}
\centering
\includegraphics[width=0.8\textwidth]{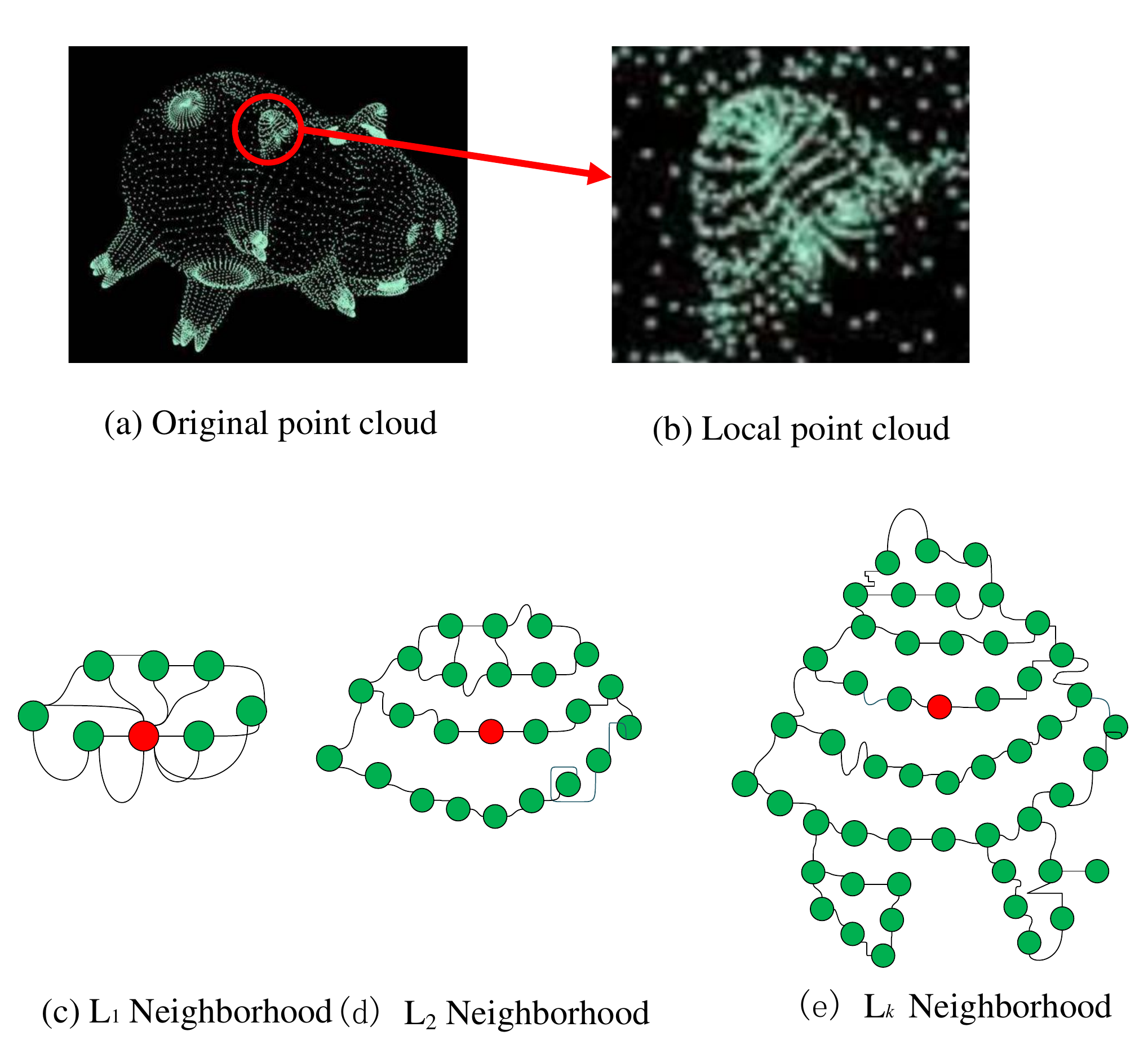}
\vspace{-0.1in}
\caption{Multi scale spatial neighborhood generation process of point cloud.}\label{fig:MVPROCESS}
\vspace{-0.2in}
\end{figure}

For an \textbf{r-scale graph}, we adopt the vertices in the point cloud directly as the nodes in the scale graph; then, we generate the edges between the vertices by constructing the multiscale neighborhoods of the vertices in the point cloud. Finally, the multiscale graphs of the point cloud are constructed. The steps for generating multiscale neighborhoods of vertices are shown in Figure \ref{fig:MVPROCESS}.

In Figure \ref{fig:MVPROCESS}, (a) is the original input point cloud; (b) is a part of the point cloud;(c) represents the neighborhood of a vertex in (b) at scale $L_1$; (d) represents the neighborhood of the vertex at scale $L_2$; (e) represents the neighborhood of the vertex at scale $L_k$; and the green point represents the neighborhood of the red vertex at each scale.
Because of the inconsistent point density, we use the average distance between points as the measuring baseline. First, we use octree \cite{OCTNET201641} method to calculate the average distance between points:
\begin{equation}
{{d}_{m}}=\sum\limits_{i=1}^{n}{\sqrt{{{(\frac{{{x}_{i}}}{ocx_{dis}})}^{2}}+{{(\frac{{{y}_{i}}}{ocx_{dis}})}^{2}}+{{(\frac{{{z}_{i}}}{ocx_{dis}})}^{2}}}}
\end{equation}

where $n$ represents the number of vertices in the point cloud, $x$, $y$ and $z$ represent the 3D coordinates of the vertices, respectively, and $ocx_{dis}$ represents the distance between octrees constructed on the point cloud. The formulas are as follows:

\begin{equation}
oc{{x}_{di{{s}_{x}}}}=\frac{\max (x)-\min (x)}{n}
\end{equation}
\begin{equation}
oc{{x}_{di{{s}_{y}}}}=\frac{\max (y)-\min (y)}{n}\\
\end{equation}
\begin{equation}
oc{{x}_{di{{s}_{z}}}}=\frac{\max (z)-\min (z)}{n}\\
\end{equation}
\begin{equation}
ocx_{dis}=\sqrt{oc{{x}_{di{{s}_{x}}}}^{2}+oc{{x}_{di{{s}_{y}}}}^{2}+oc{{x}_{di{{s}_{z}}}}^{2}}
\end{equation}

where $oc{{x}_{di{{s}_{x}}}}$, $oc{{x}_{di{{s}_{y}}}}$, and $oc{{x}_{di{{s}_{z}}}}$ represent the average distances between the vertices in the $x$, $y$, and $z$ directions, respectively; $\max(x)$, $\max(y)$, and $\max(z)$ represent the maximum values of the vertices in the point cloud in the $x$, $y$, and $z$ directions, respectively; $\min(x)$, $\min(y)$, and $\min(z)$ represent the minimum values of the vertices in the point cloud in the $x$, $y$, and $z$ directions, respectively, and $n$ represents the number of vertices in the point cloud.

\par
Then, we use the average distance to determine the scale size:
\begin{equation}
{{L}_{k}}={{2}^{k}}{{d}_{m}}
\end{equation}

where $k$ is the magnification factor, $L_k$ is the scale radius of the magnification factor, and $d_m$ is the average distance between the vertices in the point cloud. We take the magnification factor $k$ as the hyperparameter of the model and generate the neighborhood radius at different scales by selecting different $k$ values.

\par
Following the above steps, we form a multiscale graph of a point cloud represented as $G=(V,E)$. $V$ is expressed as follows:

\begin{equation}
V_{N*F}=
\left(                 
  \begin{array}{cccc}   
    f_{11} & f_{12} & $...$ & f_{1F}\\  
    f_{21} & f_{22} & $...$ & f_{2F}\\  
    $...$    & $...$    & $...$ & $...$   \\
    f_{(N-1)1} & f_{(N-2)2} & $...$ & f_{(N-1)F}\\
    f_{N1} & f_{N2} & $...$ & f_{NF}
  \end{array}
\right)                 
\end{equation}
where $N$ is the number of vertices, $F$ is the feature dimension of vertices, and $f_{ij}$ represents the $j$-th feature dimension of vertex $i$. The expression for the edges between vertices is:

 \begin{equation}
E_{N*N}=
\left(                 
  \begin{array}{cccc}   
    e_{11} & e_{12} & $...$ & e_{1N}\\  
    e_{21} & e_{22} & $...$ & e_{2N}\\  
    $...$    & $...$    & $...$ & $...$   \\
    e_{(N-1)1} & e_{(N-2)2} & $...$ & e_{(N-1)N}\\
    e_{N1} & e_{N2} & $...$ & e_{NN}
  \end{array}
\right)                 
\end{equation}

where $e_{ij}$ represents an attribute of an edge between vertices $i$ and $j$. We use Formula (11) to express $e_{ij}$:

\begin{equation}
{{e}_{ij}}=\left\{ \begin{matrix}
   ({{d}_{ij}},{{\theta }_{i}}_{j})  \\
   0  \\
\end{matrix}\begin{matrix}
   {}  \\
   {}  \\
\end{matrix} \right.\begin{matrix}
   {}  \\
   {}  \\
\end{matrix}\begin{matrix}
   if  \\
   if  \\
\end{matrix}\begin{matrix}
   {{d}_{ij}}\le {{L}_{k}}  \\
   {{d}_{ij}}\succ {{L}_{k}}  \\
\end{matrix}
\end{equation}

where $d_{ij}$ represents the distance between vertices $i$ and $j$, and $\theta_{ij}$ represents the azimuth between vertex $i$ and vertex $j$. The azimuth angle between vertex $i$ and vertex $j$ is the clockwise horizontal angle between the north direction line of point $i$ and the $ij$ line. When the distance between vertices, i.e., $d_{ij}$(calculated according to Formula (2)) is less than or equal to the scale size ($L_k$), we add an edge between the two vertices whose attribute is ($d_{ij}$,$\theta_{ij}$ ). When the distance between vertices is larger than the scale size $L_k$, no edge is added between the vertices.

By following the above steps, we can generate a corresponding multiscale point cloud graph for any point cloud data P(N), as shown in Formula 12:

\begin{equation}
P(N)=(G(L_1),G(L_2),...,G(L_k))
\end{equation}

where $P(N)$ is a multiscale graph of the point cloud with $N$ vertices, $L_k$ indicates that the scale factor of the graph is $k$, and $G(L_k)$ represents the structure of the point cloud graph with a scale of $L_k$. In this paper, when the magnification factor is set to $k$, the scale selection range of the multiscale graph is ${1,2,…,k}$. For example, when $k=3$, $P(N)=(G(L_1 ),G(L_2 ),G(L_3 ))$.

\par
Using the above steps, any point cloud can be represented by multiscale point cloud graphs. The key step in this process is determining how to select the baseline scale for a point cloud. Scales that are too large or too small cannot effectively obtain the multiscale features of the point cloud. Therefore, to meet the requirements for sparse and uneven density point cloud feature extraction, we adopt the average distance between points in the point clouds ($d_m$) as the basis of multiscale change.
\par
The features of different scale point cloud graphs can be extracted from the constructed multiscale point cloud graphs. Then, these features can be combined to enhance the descriptive ability of the local point cloud features. The graph data are unordered and their dimensions vary, which makes it difficult for traditional convolution networks to address the data in non-Euclidean space. Therefore, we propose an adaptive convolution layer for multiscale graphs that generalizes the traditional convolution operation, allowing a convolution neural network to work with irregular point cloud graph data.

\subsection{Self-adaptive graph convolution}

\subsubsection{Overview}

\begin{figure}[htbp]
\centering
\subfigure[image Convolution]{\label{fig:a}
\begin{minipage}[c]{0.5\textwidth}
\centering
\includegraphics[width=10cm]{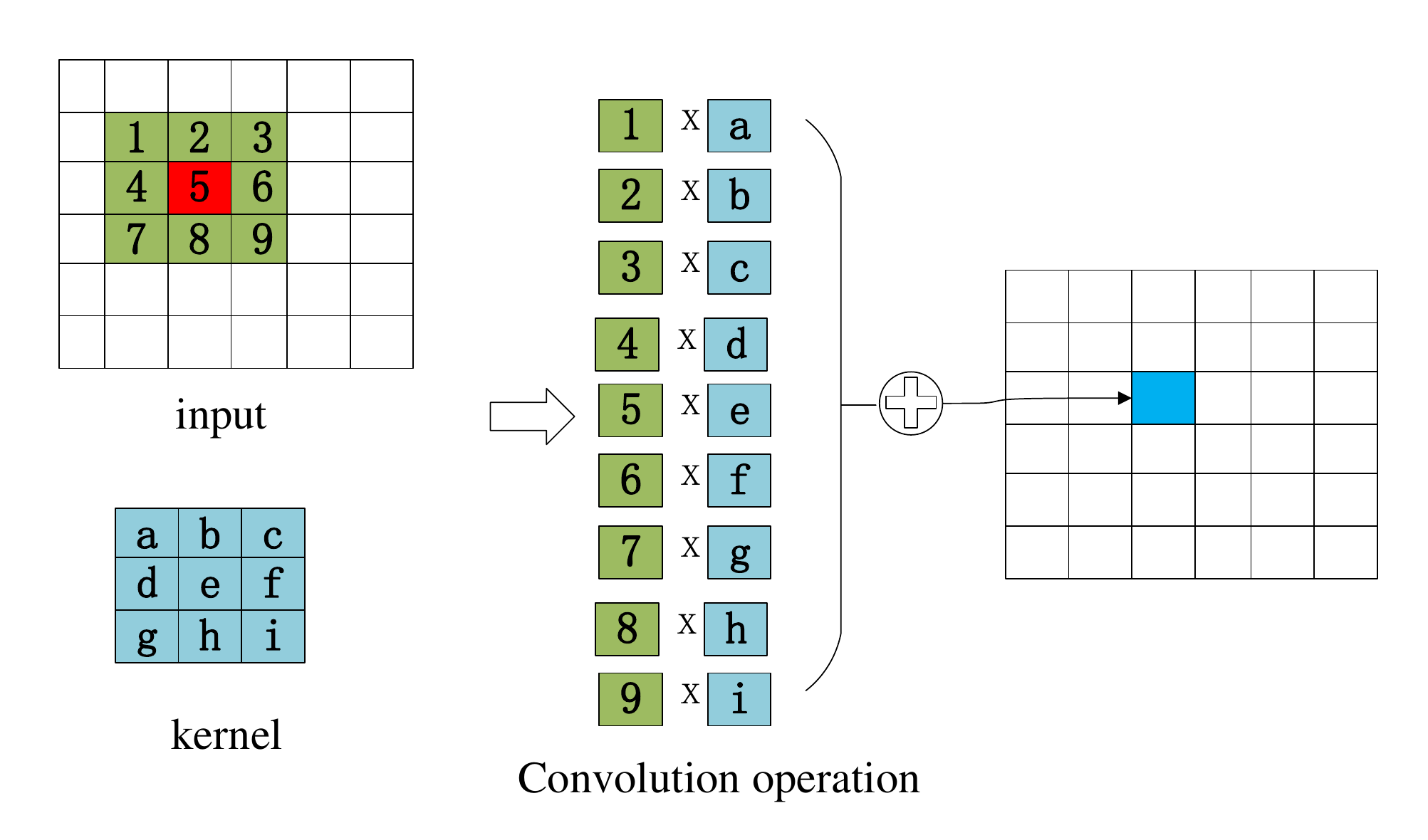}
\end{minipage}%
}%

\subfigure[Convolution of point cloud graph]{\label{fig:b}
\begin{minipage}[c]{0.5\textwidth}
\centering
\includegraphics[width=10cm]{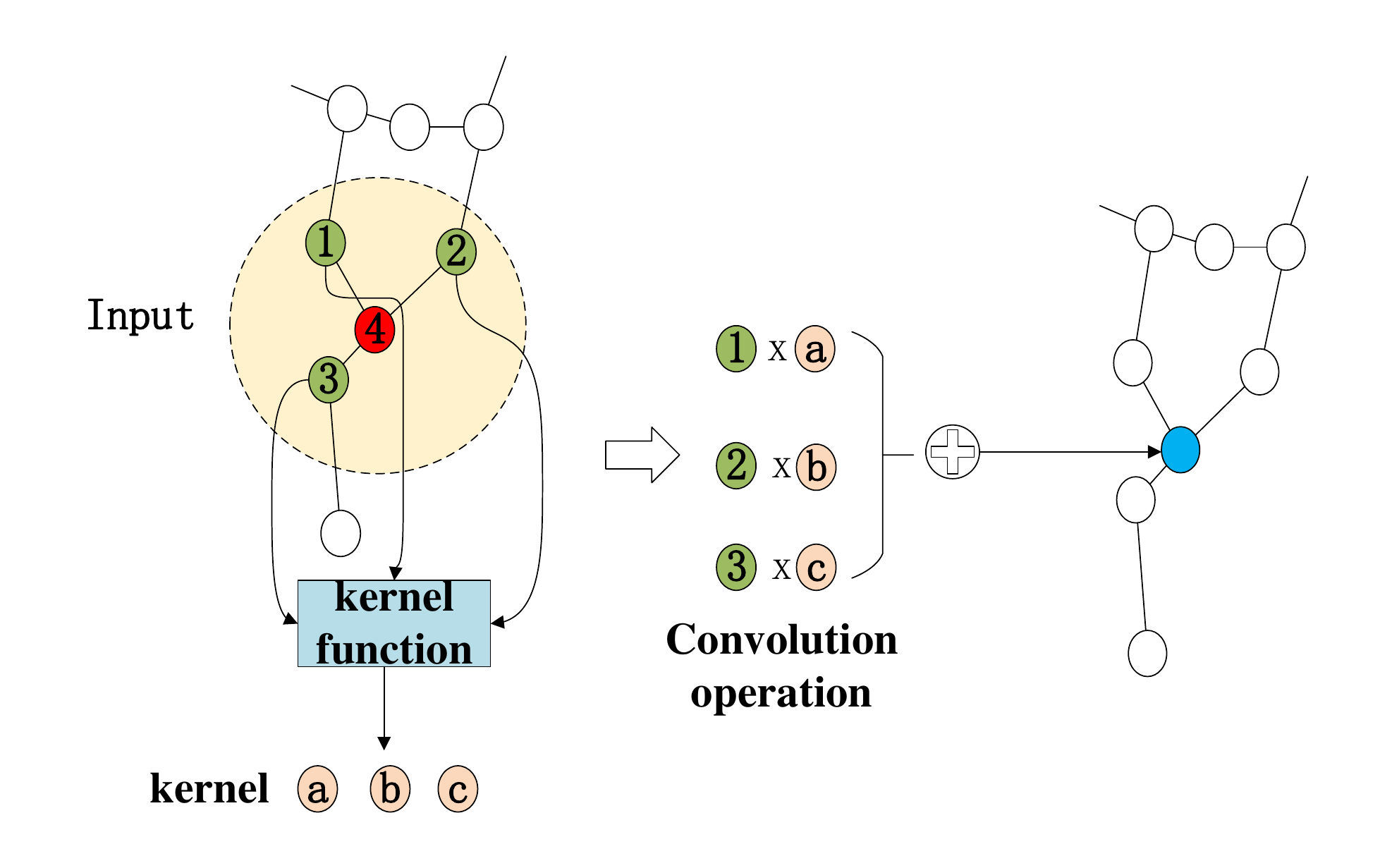}
\end{minipage}
}
\caption{Convolution of image and point cloud graph}\label{fg:convolution}
\end{figure}

Convolution is essentially an aggregation operation applied to local data. The process of learning the convolution kernel is actually the process of learning the local aggregation parameters. Each parameter can be shared. As shown in Figure \ref{fig:a}, in an image, the number and location of the neighborhood nodes of a designated node are fixed (i.e., the red node 5 is fixed); therefore, we can define a discrete convolution on node 5 and its neighborhood nodes. However, as shown in Figure \ref{fig:b}, the number and order of vertices in a point cloud graph are not fixed; thus, the traditional definition of the convolution kernel cannot achieve convolution of the point cloud graph data. Therefore, to aggregate the neighborhood information of any vertex and obtain the feature of point cloud graph data, we generalize the discrete finite-dimensional convolution kernel into a continuous infinite-dimensional convolution kernel, that is, a continuous derivable function.

\subsubsection{Point cloud graph convolution}

In traditional convolution, the convolution kernel is represented as a discrete vector. A graph with N pixels can be represented by a regular grid, such as I:$x^2 \mapsto R^n$. In this paper, we use W to express a $(2m-1)$×$(2m-1)$ convolution kernel. where m represents the size of the convolution kernel. The standard convolution operation is as follows:
\begin{equation}
{{y}_{i}}={{w}^{T}}{{x}_{i}}=\sum\limits_{{i-m}\prec j\prec {i+m}}{{{w}_{j-i+m}}*{{x}_{j}}}\\
\end{equation}
where $y_i$ represents the convolution output of the $i$-th pixel, $w$ represents the convolution kernel parameter, and $x_j$ represents the value of pixel $j$.
\par
To aggregate the local information of any vertex, we generalize the discrete finite dimensional convolution kernel into a continuous infinite dimensional convolution kernel, i.e., a univariate function $f$. By default, the relative location of each node is fixed in traditional convolution, which constitutes a strong local structure; however, this type of convolution is suitable only for regular and ordered data, such as the pixels in images. Obviously, the multiscale point cloud graph data constructed in the preceding section do not conform to this rule. Thus, to address the irregular and unordered data of the point cloud multiscale graph, we generalize this local structure description as follows:

\begin{equation}
{{y}_{i}}=\sum\limits_{{{e}_{i}}_{j}\in \xi }{f({{e}_{i}}_{j})*{{x}_{j}}}
\end{equation}

\par
where $y_i$ represents the convolution output, $e_{ij}$ represents the edge weight between the vertices $i$ and $j$, $x_j$ represents a neighborhood vertex of vertex $x_i$, and $\xi$ represents the set of edges formed by vertex i and its neighborhood vertices. Here, $f(e_{ij})$ is the convolution kernel function with $e_{ij}$ as a parameter; it is also the key component in the adaptive convolution kernel. The function f must:
(1) be continuous and derivable. Our method is based on gradient descent, i.e., the gradient to update the parameters is calculated through reverse transfer. Therefore, $f$ must be continuous and derivable.
\par
(2) have good fitting ability and generalizability.
\par
(3) have high calculation efficiency.
\par
Among the above three properties, (1) and (2) must be satisfied. Although (3) is not strictly necessary, it is important for practical applications with real-time requirements. In \cite{DGCNN201740}, the mixed Gaussian function (GMM) is used to fit the convolution kernel function, but the efficiency of this method is low. In \cite{Simonovsky2017Dynamic}, a multilayer perceptron (MLP) is used to fit the graph convolution kernel function. The hidden layers of an MLP are fully connected, which enables them to adapt to the very complex parametric model; however, at the same time, it causes the training process to be slow and have difficulty converging. Therefore, based on the above function property analysis, we design a convolution kernel of a multiscale graph based on the Chebyshev polynomial.

\subsubsection{Adaptive convolution kernel based on Chebyshev polynomials}
The optimal approximation is a theoretical method that uses a polynomial to approximate a function. Because Chebyshev polynomials can be used to construct interpolation polynomials with the best uniform approximation property, in this paper, we use Chebyshev polynomials to fit the convolution filter of the graph, which allows the graph convolution operation to run efficiently on a GPU. This efficiency occurs because the $n$-th order basis functions (n $\ge$ 1) of Chebyshev polynomials can be represented by the $n$ -1order basis functions and $n$-2 order basis functions; therefore, the function filter learning process can be decomposed into a series of iterative matrix addition operations.
First, we define the following convolution kernel function for kernel $w$:

\begin{equation}
{{f}_{w}}(d,\theta )={{g}_{w}}(d)*{{g}_{w}}(\theta )
\end{equation}
\par
where $d$ is the spatial distance between the vertices and $\theta$ represents the azimuth between vertices, both of which come from the edge attributes of multiscale point cloud graphs, and $g_w$() represents the Chebyshev function which takes the distance or azimuth between vertices as the input. The Chebyshev function has the following recursive relation:

\begin{equation}
 \begin{aligned}
 & {{T}_{0}}(x)=1 \\
 & {{T}_{1}}(x)=x \\
 & {{T}_{n+1}}(x)=2x*{{T}_{n}}(x)-{{T}_{n-1}}(x)
 \end{aligned}
\end{equation}

\par
where $x$ is the input value of the Chebyshev polynomial and $T_n$ represents the $n$-th term of the Chebyshev polynomials. From the above formula, we can see that any $T_{(n+1)}$ can be obtained by computing $T_n$ and $T_{(n-1)}$. Therefore, the Chebyshev polynomials used in this paper are:


\begin{equation}
{{g}_{w}}(x)=\sum\limits_{n=0}^{N}{{{w}_{n}}}{{T}_{n}}(x)
\end{equation}

where $w_1$,$w_2$,$……$,$w_n$ are the weights of the Chebyshev polynomials. Then, combining formulas (15), (16) and (17), the form of the convolution kernel function adopted in this paper is as follows:

\begin{equation}
{{f}_{w}}(d,\theta )=\sum\limits_{n=0}^{N}{w_{n}^{d}{{T}_{n}}(d)*\sum\limits_{n=0}^{N}{w_{n}^{\theta }{{T}_{n}}(\theta)}}
\end{equation}
where $w_n^d$ and $w_n^\theta$ represent the weights corresponding to the basis function and $N$ represents the order of the basis function in Chebyshev polynomials. The basis functions for different orders can be used to fit adaptive convolution kernels of different complexity. In this paper, $N$ is taken as the model hyperparameter, and the appropriate order is selected through experimental verification.

\par
Because the gradient descent algorithm and its variant SGD affect the efficiency and accuracy of network training, this paper uses the Adam optimization algorithm during the training process. Adam was first proposed by Kingma et al.\cite{Adam2015}; it is a first-order optimization algorithm instead of the traditional SGD process and can update the neural network weights iteratively based on the training data. In this paper, the parameters are updated by the following gradient formula:

\begin{equation}
\frac{\partial {{f}_{w}}(d,\theta )}{\partial w}=\sum\nolimits_{n\in (0,N)}{\frac{\partial {{g}_{w}}(d)}{\partial w_{n}^{d}}*\frac{\partial {{g}_{w}}(\theta )}{\partial w_{n}^{\theta }}}
\end{equation}

where $f_w$(d,$\theta$) is the convolution kernel function of the adaptive graph with the parameters $d$ and $\theta$ as input, and $N$ is the number of basis functions of the Chebyshev polynomials.

\subsubsection{Time complexity analysis}
In this paper, matrix iteration is used to calculate the Chebyshev polynomials. At the same time, the complexity of iteratively calculating the algebraic polynomials is simplified by multiplication of the coefficient matrix. The time complexity is $O(kn)$, where $k$ represents the number of feature extraction modules. When $k$ is fixed, the time complexity of our method increases linearly with the number of vertices. When parallel computing is used, the time complexity of this method can be reduced to $O(n)$. However, the time complexities of voxel-based methods such as Subvolume \cite{GDVM201636} increase exponentially as the volume increases, and the time complexities of multiwindow-based methods such as MVCNN \cite{MVCNN2015119} increase exponentially as the image resolution increases.

\subsection{Deep neural network model based on a multiscale graph and adaptive graph convolution}
By combining the above multiscale graph and adaptive graph convolution method, we construct a multiscale feature generation module that acts on a point cloud, as shown in Figure 4:

\par
\begin{figure}
\vspace{-0.1in}
\centering
\includegraphics[width=1.05\textwidth]{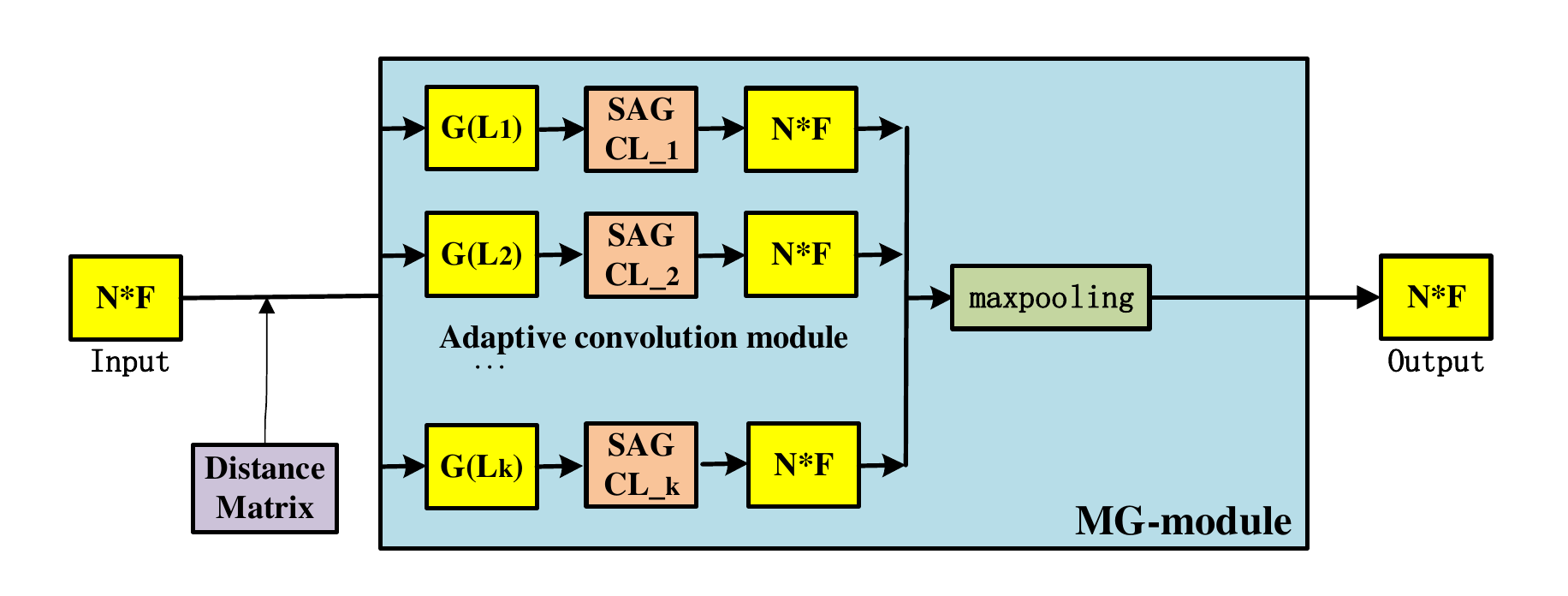}
\vspace{-0.1in}
\caption{Point cloud feature generation module of multi-scale graph.}\label{fig:PCFGM}
\vspace{-0.2in}
\end{figure}

In Figure \ref{fig:PCFGM}, the number of vertices in the point cloud input into the model is $N$, and the feature dimension is $F$. The Distance Matrix module provides the distance matrix between vertices in the multiscale graph. Then, according to the method described in Section 3.1, the input point cloud data are expressed by the multiscale graph $G(L_k)$, where $k$ is the scale size. Then, the self-adaptive graph convolution (SAGC) proposed in Section 3.2 is used to generate the local features whose dimensions are $F$ for each scale graph. Finally, the multiscale graph convolution features are fused by the max pooling module.

\par
\begin{figure}
\vspace{-0.1in}
\centering
\includegraphics[width=0.8\textwidth]{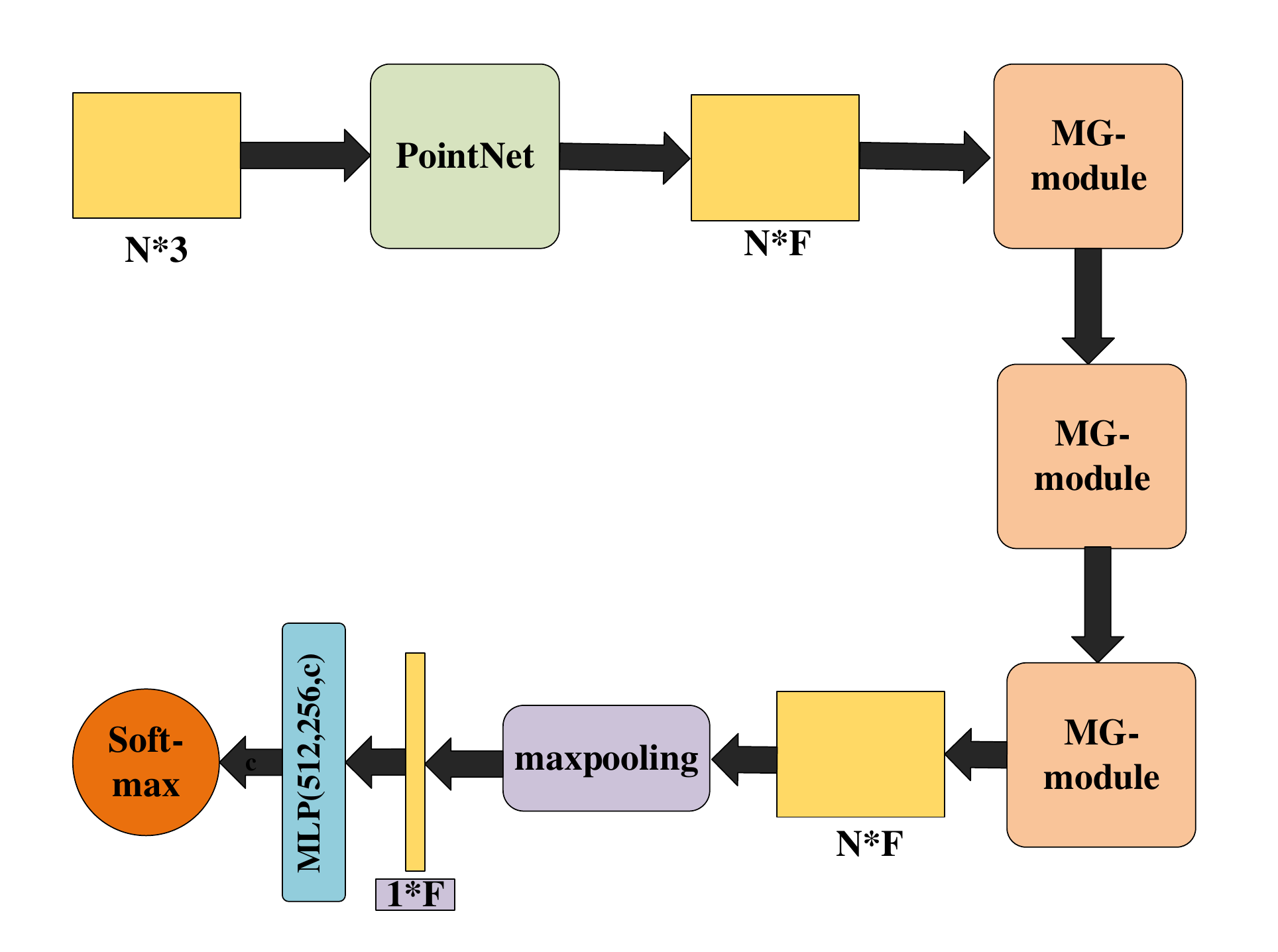}
\vspace{-0.1in}
\caption{Network framework of point cloud classification based on adaptive graph convolution.}\label{fig:NFPCL}
\vspace{-0.2in}
\end{figure}

We then append the multiscale point cloud feature generation module to the deep network structure, as shown in Figure \ref{fig:NFPCL}.

\par
In the experiments, for any 3D point cloud, we use a random sampling method to select 1,024 points as the input point cloud data and set the batch size to 32. Before performing multiscale feature learning of the point cloud, we first use PointNet \cite{PointNet20171007} to extract features whose dimensions are $F$ from each point in the point cloud. Then, we use three MG-Modules to update the features of each vertex, leaving the feature dimensions of the vertices unchanged. Then, we use the max pooling operation to obtain the global features of the point cloud. Finally, the global feature is input to the MLP and a softmax layer is used for classification.

\section{Experiments}
In the experiments, we tested the proposed MG module on two tasks: point cloud classification and point cloud retrieval. First, the experimental verification method is used to determine the hyperparameters in the MG module; then, we compare our method with other state-of-the-art methods.

\par
The parameter settings for the proposed method in this paper is unified as follows: all the models are implemented using PyTorch, on a hardware platform equipped with an M4000 GPU. We use Adam for model optimization, with the initial learning rate set to $10^-3$. To prevent overfitting, a dropout algorithm with a dropout rate set to 0.5 is added during the training process. After each convolution operation, we use batch normalization to normalize the output.

\subsection{Datasets}
We used the following datasets for testing.
\begin{itemize}
 \item ModelNet10 \cite{PPFNG2018}: ModelNet10 contains 4,899 3D point cloud model data. The number of categories is 10, of which the number of training set data is 3,991 and the number of test set data is 909.
 \item ModelNet40 \cite{PPFNG2018}: ModelNet40 contains ,311 three-dimensional point cloud model data in 40 categories. There are 9,843 data in the training set and 2,468 data in the test set.
 \item ShapeNetCore dataset \cite{POINTSIFT200946}: ShapeNetCore is a subset of ShapeNet \cite{FPFH20091007}. This dataset contains 51,300 three-dimensional point cloud data in 55 categories. The training/validation/test ratio of this dataset is 7:1:2.
\end{itemize}

\begin{figure}[htbp]
\centering
\subfigure[modelNet10]{\label{fig:precision}
\begin{minipage}[c]{0.5\textwidth}
\centering
\includegraphics[width=5cm]{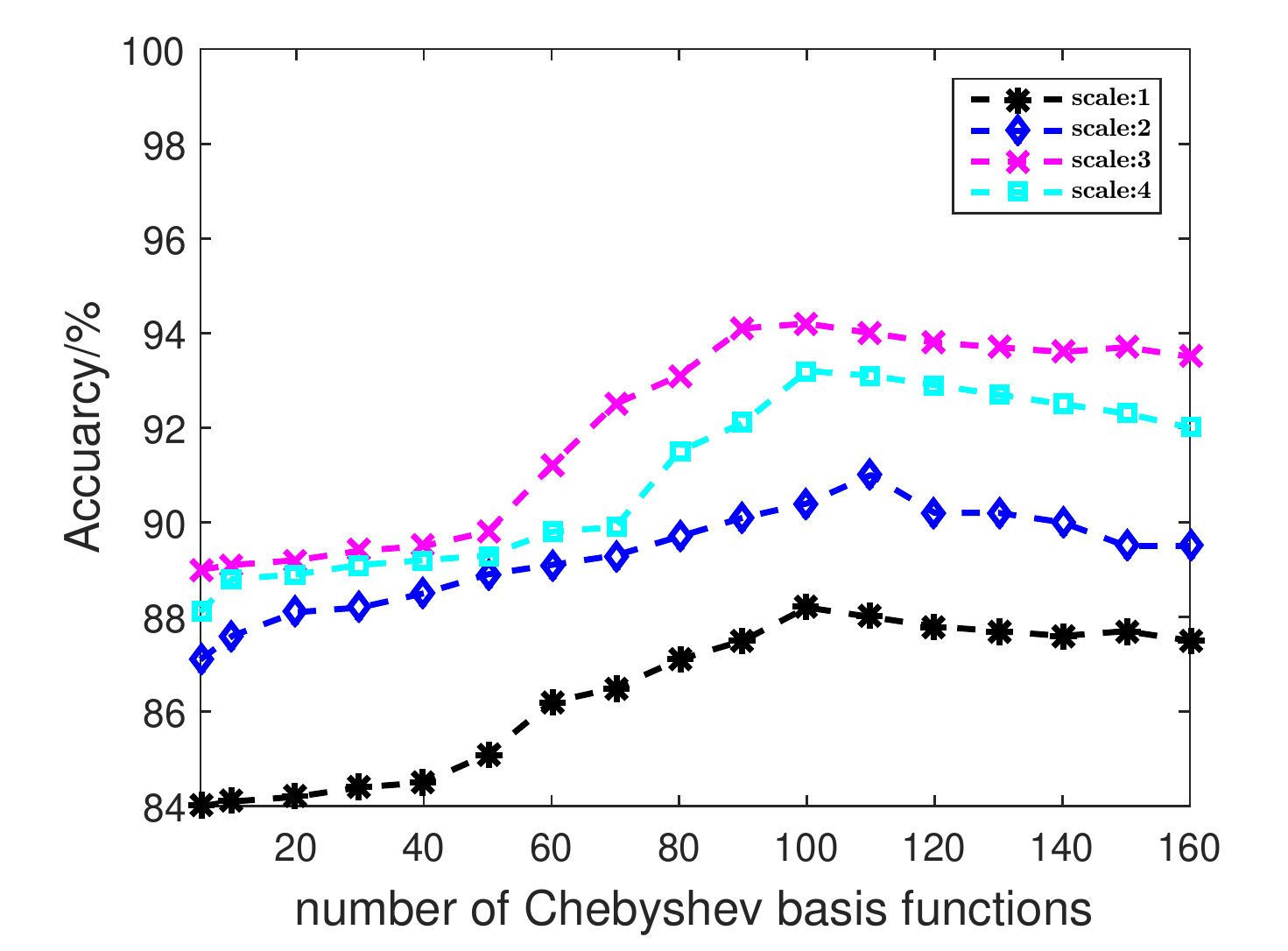}
\end{minipage}%
}%
\subfigure[modelNet40]{
\begin{minipage}[c]{0.5\textwidth}
\centering
\includegraphics[width=5cm]{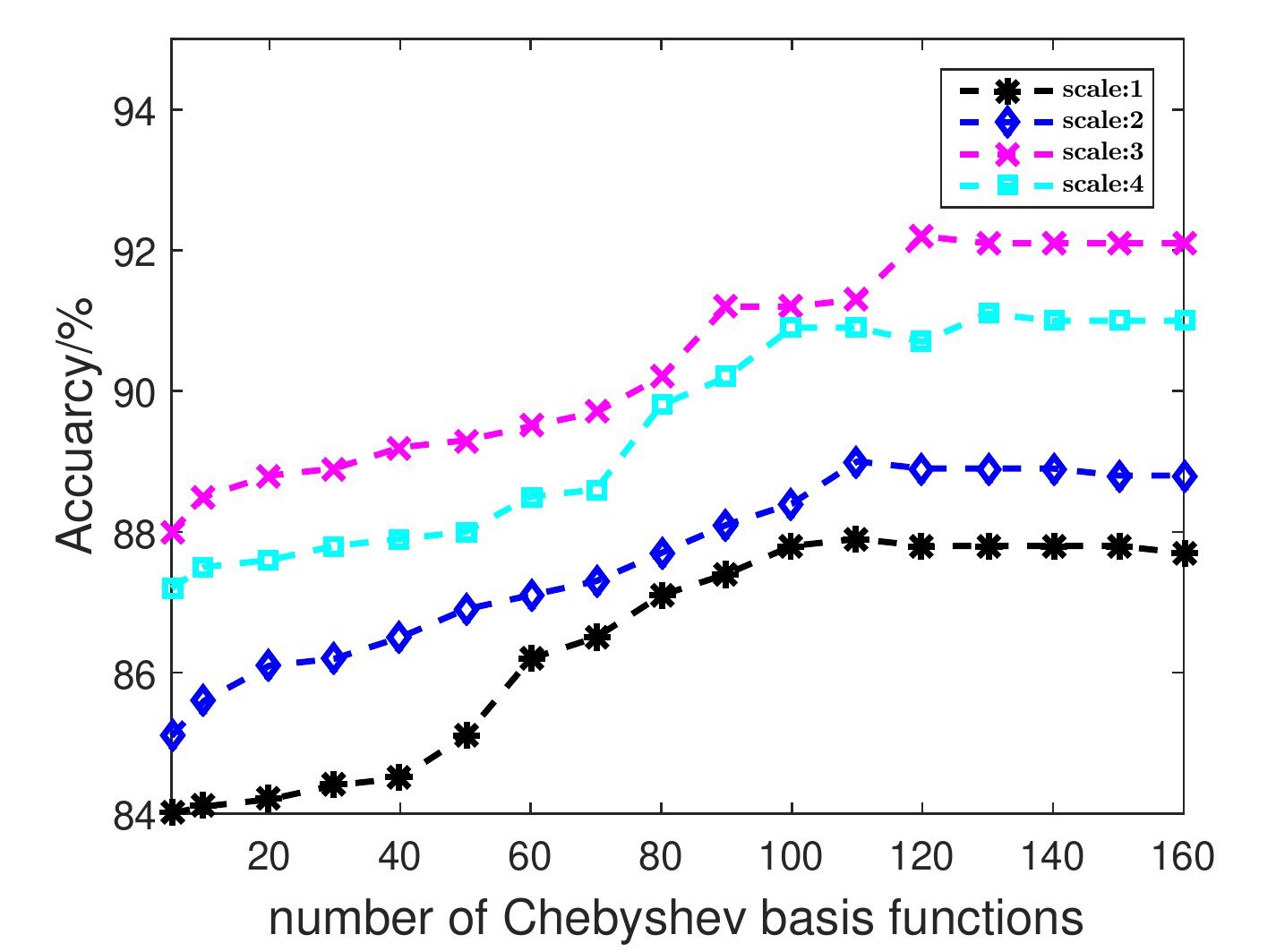}
\end{minipage}
}
\subfigure[ShapeNetCore]{
\begin{minipage}[c]{0.5\textwidth}
\centering
\includegraphics[width=5cm]{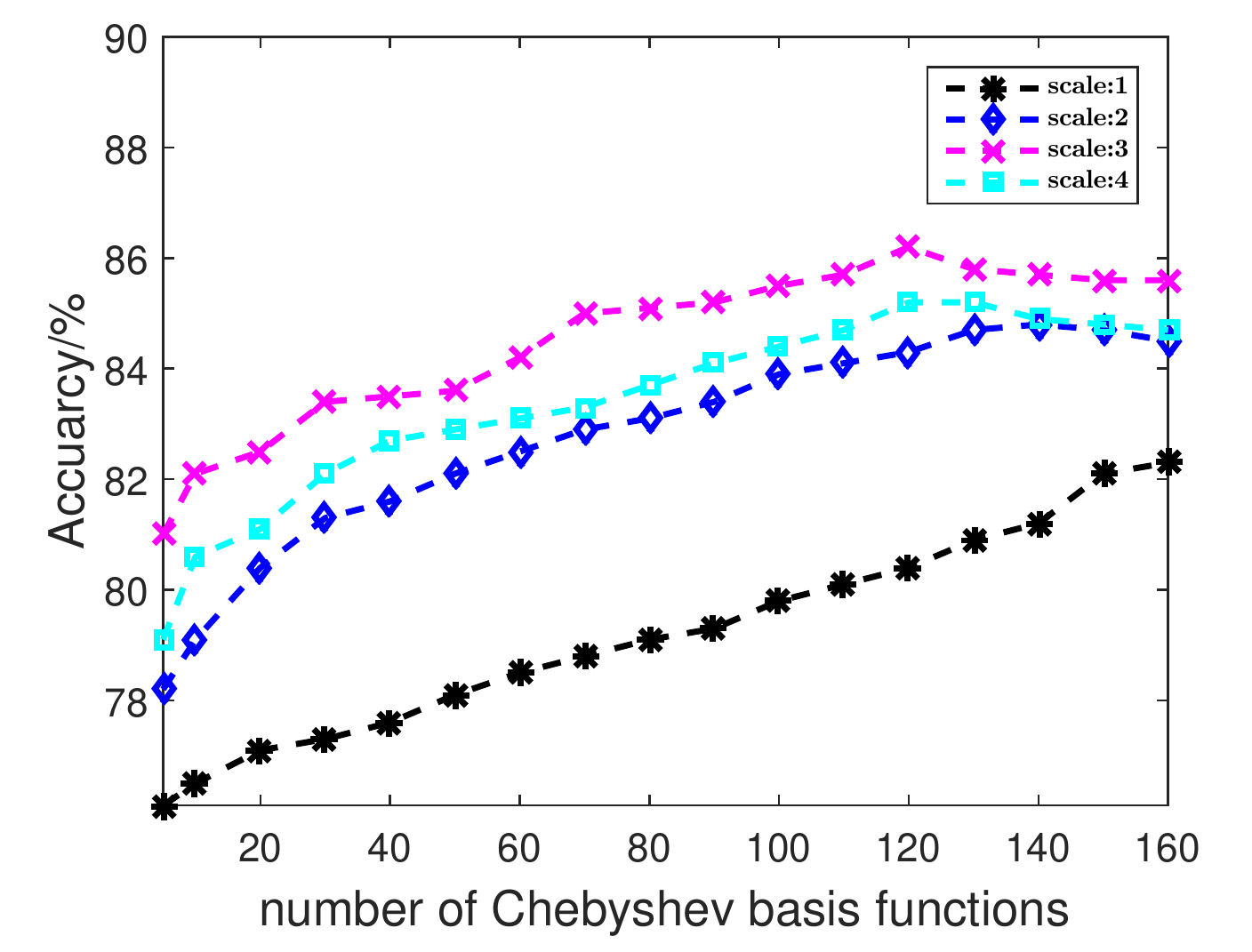}
\end{minipage}
}
\caption{The effect of the number of Chebyshev base functions and their scale range on the classification results.}\label{fg:precision}
\end{figure}

\par

\subsection{3D Point cloud classification}
Two important parameters must be analyzed in our classification network model: (1) the number of Chebyshev basis functions and (2) the scale range for constructing the multiscale graph. As shown in Figure 5, on the ModelNet10 dataset, the best effect occurs when the number of Chebyshev base functions is 100 and the scale range is 3. On the ModelNet40 and ShapeNetCore datasets, the best classification effect occurs when the number is 120 and the multiscale range is 3.

\begin {table*}[!htbp]

\caption{\label{tab:gre} the comparison results of our method with other methods}
\centering
\begin{tabular}{|c|c|c|c|}
\hline
\textit Datasets & ModelNet10 & ModelNet40 & ShapeNetCore  \\
\hline Subvolume & 90.1 & 89.2 & 85.8 \\
\hline VRN Single & 93.6	& 91.3	& 83.7 \\
\hline OctNet & 90.9	& 86.5	& 76.1 \\
\hline ECC & 90.8	& 87.4	& 81.3 \\
\hline MVCNN & 94.2 & 92.0 &	87.2\\
\hline Kd-Network(depth 15) & 94.0 & 91.8 & 83.6 \\
\hline DGCNN & 92.1 & 89.3 & 85.2\\
\hline PointNet & 91.2 & 89.2 & 84.1 \\
\hline PointNet++ & 93.1	& 91.9 &	85.3 \\
\hline PointNet + 1*MSG-net & 90.4 & 87.5 & 80.5\\
\hline PointNet + 2*MSG-net & 92.6 & 91.3 & 82.3\\
\hline PointNet + 3*MSG-net & 94.3 & 92.5 & 86.6\\
\hline PointNet + 4*MSG-net & 94.0 & 92.2 & 86.2\\
\hline PointNet + 5*MSG-net & 93.9 & 92.1 & 86.0\\
\hline
\end{tabular}
\vspace{-0.2in}
\end{table*}

\par
Following the experimental verification step described above, we selected the optimal hyperparameters for three different datasets: for ModelNet10, the number of basis functions is 100, the multiscale range parameter is 3; for ModelNet40 and ShapeNetCore, the number of basis functions is 120, and the multiscale range parameter is 3. Then, we compared our method with other state-of-the-art methods. The comparison results in Table 1 shows that when using one MG module, the classification accuracy rate of our proposed model on ModelNet10 is 90.4, on ModelNet40 it is 87.5, and on ShapeNetCore it is 80.5. As the number of MG modules increases, the classification accuracy of our model improves; however, after the number of modules reaches 3, the classification accuracy of the model no longer improves. With 3 MG modules, the classification accuracy rates of our method on ModelNet10, ModelNet40, and ShapeNetCore are 94.3, 92.5, 86.6, respectively. The experimental results show that the method proposed in this paper outperforms the other methods on the ModelNet10 and ModelNet40 datasets with regard to classification accuracy. Our method's classification accuracy on the ShapeNetCore dataset is second best and only 0.6 less than that of MVCNN.

\paragraph{Model efficiency analysis:} We randomly selected 50 points of cloud data as input from the ModelNet10, ModelNet40, and ShapeNetCore datasets and recorded the average forward-pass time, i.e., the classification time. We compared our method with the other methods that also dynamically generate convolution kernels. The comparison results are shown in Table \ref{tab:time}.

\begin {table*}[!htbp]

\caption{\label{tab:time} Comparison of the time cost between our method and other dynamic convolution kernel generation methods}
\centering
\begin{tabular}{|c|c|}
\hline
\textit Method & time(unit: ms)  \\
\hline ECC & 215.36 \\
\hline DGCNN & 823.14 \\
\hline Our method & 120.15 \\

\hline
\end{tabular}
\vspace{-0.2in}
\end{table*}

\par
As shown in Table \ref{tab:time}, when the model structure is selected as PointNet + 3 * MG-module, the forward-pass time of the entire network is 120.15 ms, which is far less than the other methods in terms of execution time. ECC uses multiple multilayer perceptrons (MLPs) to fit the convolution filter; thus, its time complexity depends on the number of hidden neurons in the MLPs, which have a fully connected structure. Therefore, ECC's time overhead is higher than that of the method proposed in this paper. The DGCNN requires a mixed Gaussian function calculation for all the vertices of each neighborhood node to obtain the convolution filter parameters; consequently, its forward-pass time is much longer than that of the method proposed in this paper.

\begin {table*}[!htbp]

\caption{\label{tab:retrieval} The comparison between our method and other methods}
\centering
\begin{tabular}{|c|c|c|c|}
\hline
\textit Datasets & ModelNet10 & ModelNet40 & ShapeNetCore  \\
\hline ECC &	91.1 & 87.0 & 81.5\\
\hline VRN Single & 92.2 & 89.3 &	73.4\\
\hline OctNet & 90.8 & 86.2 & 81.2 \\
\hline Kd-Network(depth 15) & 88.1 & 84.6 & 61.7 \\
\hline PointNet + 1*MG-module & 87.1 & 85.1 & 81.2\\
\hline PointNet + 2*MG-module & 89.2 & 87.2 & 83.1\\
\hline PointNet + 3*MG-module & 92.4 & 89.4 & 84.1\\
\hline PointNet + 4*MG-module & 92.2 & 89.2 & 84.0\\
\hline PointNet + 5*MG-module & 92.1 & 89.0 & 83.8\\
\hline
\end{tabular}
\vspace{-0.2in}
\end{table*}

\subsection{3D Point cloud retrieval}
To conduct point cloud retrieval, we first train the three classification networks on each of the three datasets and then extract the output of the 512-dimensional fully connected layer as the descriptive feature of the three-dimensional point cloud. Finally, the feature vectors are used to measure the similarity between different 3D point cloud data. The metric used to evaluate the retrieval results is the mean average precision (mAP). Table \ref{tab:retrieval} shows the comparison results.

\par
\begin{figure}
\vspace{-0.1in}
\centering
\includegraphics[width=0.8\textwidth]{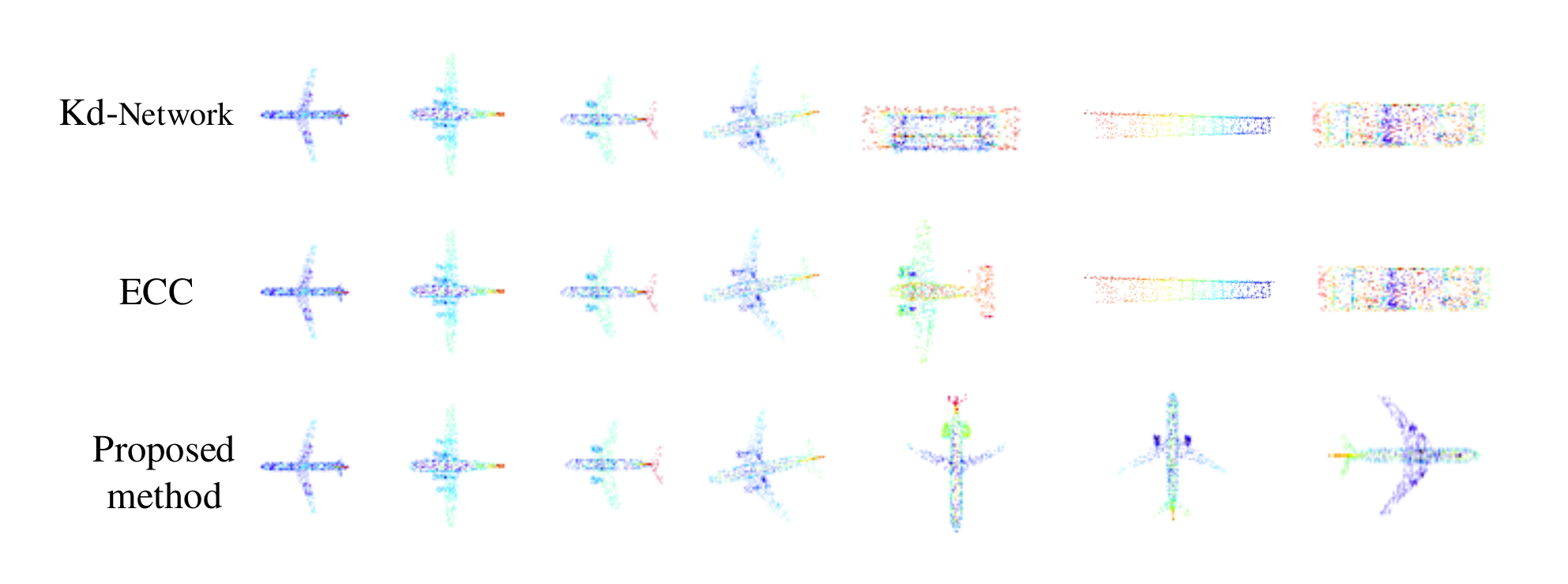}
\vspace{-0.1in}
\caption{the partial search results of our method, Kd-Network, and ECC on the ModelNet40 dataset. Each row contains the top6 search results; the object in the first column is the query.}\label{fig:Visualizing}
\vspace{-0.2in}
\end{figure}

As shown in Table \ref{tab:retrieval}, the mAPs of our method on the three datasets are 87.1$\%$, 85.1$\%$, and 81.2$\%$   when using one MG module. The retrieval accuracy of the model improves as the number of MG modules increases. Our method reaches its maximum mAP on the three datasets when three MG modules are used. When the number of modules exceeds 3, the retrieval accuracy of the model no longer improves and can even decrease. One possible reason is that after the number of MG modules reaches a certain level, the model complexity becomes too high, and overfitting is prone to occur. The experimental results show that our method is better than that of other point cloud retrieval methods on the three datasets. Figure \ref{fig:Visualizing} shows a portion of the methods' visual retrieval results. We can see that the retrieval effect of our method is better than others, especially when the objects are rotated.

\par
Figure \ref{fig:Visualizing} visually shows the partial search results of our method, Kd-Network, and ECC on the ModelNet40 dataset. Each row contains the top6 search results; the object in the first column is the query.

\section{Conclusions}
This paper presents a new graph convolution network based on a multiscale point cloud graph and an adaptive graph convolution kernel. First, we propose the method for constructing multiscale graphs of 3D point clouds; then, we present the convolution kernel for the adaptive graph based on the Chebyshev polynomial to dynamically generate the convolution parameters. The adaptive graph convolution kernel solves the problem of implementing convolution operations on disordered graph vertex neighborhoods. We conducted classification and retrieval experiments on the Modelnet10/40 and ShapeNetCore datasets and analyzed the hyperparameters, such as the number of different scales and Chebyshev base functions, through experiments. Compared with other point cloud deep neural network algorithms, our method achieves better classification and retrieval results is computationally more efficient. Future work may include testing different convolution kernel generation functions to attempt to further improve the effect and efficiency of point cloud classification and retrieval.

\section*{References}
\bibliography{mybibfile}

\end{document}